\documentclass{article}
\usepackage{amsmath,graphicx}
\usepackage{amssymb}
\usepackage{booktabs}
\usepackage[a4paper, left=1in, right=1in, top=1.5in, bottom=1.5in]{geometry}

\title{Shifts in Doctors' Eye Movements Between Real and AI-Generated Medical Images}
\author{David C Wong$^{1}$, Bin Wang$^{1}$, Gorkem Durak$^{1}$, Marouane Tliba$^{2}$,\\ Mohamed Amine Kerkouri$^{3}$, Aladine Chetouani$^{4}$, Ahmet Enis Cetin$^{5}$,\\ Cagdas Topel$^{1}$, Nicolo Gennaro$^{1}$, Camila Vendrami$^{1}$,\\ Tugce Agirlar Trabzonlu$^{1}$, Amir Ali Rahsepar$^{1}$, Laetitia Perronne$^{1}$,\\ Matthew Antalek$^{1}$, Onural Ozturk$^{1}$, Gokcan Okur$^{1}$,\\ Andrew C. Gordon$^{1}$, Ayis Pyrros$^{6}$, Frank H Miller$^{1}$,\\ Amir A Borhani$^{1}$, Hatice Savas$^{1}$, Eric M. Hart$^{1}$, Elizabeth A Krupinski$^{7}$, Ulas Bagci$^{1}$
\\ \\
$^{1}$Northwestern University, Evanston, Illinois, USA\\
$^{2}$University of Orleans, Orleans, France\\
$^{3}$f-initiatives, Paris, France\\
$^{4}$University Sorbonne Paris Nord, Villetaneuse, Paris, France\\
$^{5}$University of Illinois Chicago, Chicago, Illinois, USA\\
$^{6}$Duly Health and Care, Chicago, Illinois, USA\\
$^{7}$Radiology and Imaging Science, Emory university, Atlanta, Georgia, USA 
}

\date{}

\begin{document}
%
\maketitle

\begin{abstract}
Eye-tracking analysis plays a vital role in medical imaging, providing key insights into how radiologists visually interpret and diagnose clinical cases. 
In this work, we first analyze radiologists’ attention and agreement by measuring the distribution of various eye-movement patterns, including saccades direction, amplitude, and their joint distribution. These metrics help uncover patterns in attention allocation and diagnostic strategies. Furthermore, we investigate whether and how doctors’ gaze behavior shifts when viewing authentic (Real) versus deep-learning-generated (Fake) images. To achieve this, we examine fixation bias maps, focusing on first, last, short, and longest fixations independently, along with detailed saccades patterns, to quantify differences in gaze distribution and visual saliency between authentic and synthetic images.
\end{abstract}
\footnotetext{This Paper was accepted at ETRA'25 GenAI Workshop, Tokyo, Japan.}
\section{Introduction}

Medical imaging is fundamental to disease diagnosis and treatment planning \cite{Bercovich2018}. Techniques like radiography, computed tomography (CT), magnetic resonance imaging (MRI), and positron emission tomography (PET) enable visualization of internal structures and the detection of abnormalities \cite{Atabo2019}. Radiologists play a critical role in interpreting these images, identifying lesions, tumors, and other anomalies to ensure accurate diagnoses \cite{Hussain2022}.

Recent advancements in AI, particularly Deep Neural Networks, have significantly improved the analysis and interpretation of medical images, enhancing diagnostic accuracy and efficiency \cite{Najjar2023, rad2023}. Generative AI techniques, such as Generative Adversarial Networks (GANs) \cite{GANs} and diffusion-based models \cite{stable_diffusion}, have facilitated the creation of synthetic medical images. These synthetic images are increasingly used to augment training datasets, addressing data scarcity and improving diagnostic precision, especially when real images are limited or of poor quality \cite{synt_med_gen1, synt_med_gen2, Islam2024, brain_img_syth}.
Despite their potential, AI-generated medical images raise ethical and regulatory concerns about their quality, viability, and reliability \cite{Musalamadugu2023}. Ensuring these images are both AI-recognizable and clinically realistic for radiologists is crucial for their safe and effective integration into medical workflows \cite{Kelkar2023}.

Eye-tracking technology is instrumental in analyzing radiologists' visual and cognitive behavior \cite{ETML_med,Trkan2007HumanEL}. By tracking scanpaths—the sequence of gaze movements across a stimulus—it identifies fixation points, highlighting regions that capture attention during image interpretation \cite{Kerkouri2022, Tliba_2022_CVPR, 1334654}. Studying these visual patterns enhances the understanding of diagnostic strategies and AI interpretability \cite{Wu2019, ETML_med}. Research shows that experienced radiologists exhibit more efficient visual search behaviors than novices, due to their expertise in recognizing critical image features \cite{Lou2021}. Additionally, eye-tracking data can improve AI models by integrating radiologists' visual attention into machine learning algorithms, enabling AI to better simulate expert decision-making \cite{Leveque2021, K2024}.

Eye-tracking technology also serves as a tool for performance assessment and targeted training in medical imaging \cite{Pershin2023}. However, the increasing use of image generative models \cite{GANs, stable_diffusion, 9051780} in medical imaging raises several regulatory and ethical concerns. These technologies can be misused to provide false diagnostics, fake exam results, or false treatment plans, leading to adverse outcomes for patients \cite{Zingarini2024}. When generated images exhibit high realism, they may become challenging to differentiate, even for expert radiologists.

To the best of our knowledge, no prior research has explored how AI-generated medical images influence the eye movements of radiology experts. Understanding shifts in visual attention mechanisms can provide critical insights into cognitive processes, improve radiologist training, and enhance diagnostic accuracy. Additionally, it can aid in refining generative models for medical imaging, ensuring greater reliability and clinical applicability. This research also contributes to the development of ethical guidelines for integrating AI-generated images into medical practice while safeguarding patient trust and safety.
This study bridges the gap by analyzing statistical variations and behavioral patterns in radiologists' gaze distribution when interpreting real versus AI-generated chest X-ray images. Our key contributions include:
\begin{itemize}
\item \textbf{Dataset Development}: Creation of a dataset combining real and AI-generated chest X-ray images across diverse clinical conditions.
\item \textbf{Eye-Tracking Experimentation}: Design and implementation of eye-tracking studies with expert radiologists to assess their ability to differentiate AI-generated from authentic X-rays.
\item \textbf{Gaze Pattern Analysis}: In-depth statistical evaluation of radiologists’ visual attention to identify patterns and variations when interpreting real versus synthetic images.
\item \textbf{Behavioral Insights}: Demonstration of significant differences in visual search strategies, highlighting the cognitive impact of AI-generated images on medical decision-making.
\end{itemize}
This work provides a foundation for improving AI-assisted radiology, optimizing medical image generation, and guiding future regulatory frameworks for AI integration in clinical practice.

\section{Related Works} 
Eye tracking 
Von Helmholtz's pioneering research  In his \textit{Treatise on Physiological Optics} \cite{helmholtz1924treatise}, elucidated visual attention as a fundamental mechanism in visual perception. The seminal experimental work by Yarbus \cite{Yarbus1967} involved tracking subjects' eye movements while they viewed images and responded to questions, thereby ensuring the monitoring of their attention.
While Kosslyn's work \cite{Kosslyn1994} describes attention as a selective process within perceptual processing, facilitated by an 'attentional window' that filters information in the 'visual buffer.' This adaptable window modulates its transmission capacity, allowing certain information to be processed downstream while excluding other data.

Eye Tracking and medical images The work of \cite{Nodine1987} was one of the first works that explored eye tracking for professional clinical application of improving tumor detection. Later, the work of \cite{KUNDEL1990} concluded that providing visual feedback based on eye-position and gaze-duration times significantly improves the detection of nodules in chest radiographs. Specifically, radiology residents who received visual feedback showed an average improvement of $16\%$ in their performance.

The study by \cite{Litchfield2010} demonstrates that radiographers' diagnostic performance improves when they observe the search behavior of others, regardless of the observer's expertise. This improvement is particularly significant when the observed eye movements are related to searching for nodules, with novices showing consistent improvement when following an expert's search behavior.

More recent studies like \cite{Borg2018PreliminaryEU} explored whether eye tracking could distinguish novices from experts in interpreting ultrasound-guided regional anesthesia images. It found that while both groups had similar gaze times in the area of interest (AOI), novices spent more time outside the AOI and took longer to answer and fixate on the AOI. The study concluded that experts are quicker at identifying sonoanatomy and spend less unfocused time, suggesting that eye tracking is an effective tool for differentiating levels of expertise in ultrasound image interpretation.
\cite{Leveque2021}, studied the Gaze Patterns of Expert Radiologists in Screening Mammography for cancer detection, which revealed that there is a significant difference in visual behaviour between radiologists working in  the same environment depending on their degrees of experience.

AI Image Generation 
Advancements in generative AI models such as GANs \cite{GANs}, Stable Diffusion \cite{stable_diffusion}, and Roentgen \cite{Bluethgen2024} have expanded their applications to medical imaging, addressing data scarcity and enhancing training for diagnostic models.    A GAN-based approach  was introduced in \cite{Nie2018}  to synthesize medical images, reducing acquisition costs while preserving realism through adversarial learning. The method proposed by \cite{Lee2020} generated synthetic MRI images from CT scans using GANs, demonstrating high similarity to real images.  A multi-stage GAN framework was developed by \cite{Ciano2021} for generating X-ray images with segmentation labels, facilitating data augmentation for AI-based diagnosis.  Diffusion models have also been explored for medical image synthesis.  SegGuidedDiff was introduced  in \cite{diff_anat}, leveraging segmentation masks for anatomically controlled image generation.  the potential of diffusion models for generating MRI and CT image synthesis was demonstrated in \cite{med_diff}  , improving segmentation tasks through self-supervised learning.  

Eye-Tracking in AI-Generated Content  
Research on AI-generated images and eye movement patterns remains limited. Studies in misinformation detection \cite{Steinfeld2023} show that digital literacy influences scan patterns, with users focusing on credibility cues. Similarly, \cite{huang2024analysishumanperceptiondistinguishing} examined human perception of AI-generated faces, showing increased scrutiny of images suspected to be fake.  While eye-tracking has been used to analyze deepfake detection, no prior work has studied its effects in medical imaging. This study addresses this gap by investigating how AI-generated medical images influence radiologists' gaze behavior and decision-making.

\section{Dataset}
\subsection{Experimental Protocol}

Previous works demonstrated that gaze patterns differ based on image intrinsic features as well as the users' cognitive profile. 
We aim to understand how experts perceive and interact with synthetic data through a perceptual test, with the hypothesis : The gaze patterns of radiologists change when viewing AI-Generated X-Ray images vs real X-Ray images.

The participating radiologists were briefed on the tasks they would be performing and the data collected, with the objective of classifying the images into real and fake and making a diagnostic decision.  After the experiment, the radiologists were briefed on whether the chest X-rays were real or generated, as well as the analyses that would be performed. 
This study takes place in a dark environment designed to replicate a normal radiology reading room environment. To track eye gazes, we used the EyeLink 1000 Plus device, which uses a single camera at 500 Hz to track eye movements for both pupils. The eye tracker is set 30cm in front of the monitor, and the radiologist begins by sitting with their face approximately 80cm away from the center of the monitor \cite{cornelissen2002eyelink}. Gaze points are determined as the average of the two pupils with fixations being defined as an area where saccades have a velocity of less than 30°/s and acceleration less than $8000^\circ/s^2$ \cite{cornelissen2002eyelink}.
The eye tracker is calibrated through a 13-point calibration process before each experiment, which was validated within the EyeLink calibration system \cite{cornelissen2002eyelink}. Throughout the experiment, the system is also corrects blinking, head movements, and other artifacts. Images were presented in  a randomized order, where each participant conducted a single trial.

\subsection{Analysis of Stimuli}
\begin{figure}[h!]
    \centering
    \includegraphics[width=1.0\columnwidth]{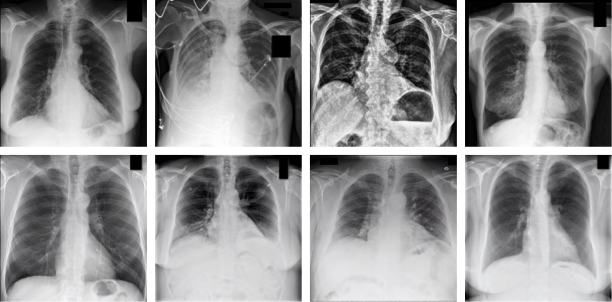} 
    \caption{The first row displays real X-ray images, while the second row shows generated images. The diagnostic classifications of the real images are: Normal, Pleural Effusion, Atelectasis, and Hyperinflated Lungs. Black boxes are used for anonymization.} 
    \label{fig:real_fake_samples}
\end{figure}

\subsubsection{Fake Stimuli}

To generate synthetic chest X-rays, RoentGen, an LDM-based generative model, is used \cite{chambon2022roentgen}. A noisy image vector is conditionally denoised by a U-Net based on the text embedding generated by the encoder. The variational autoencoder then decodes the latent vector and maps it to a pixel space, which results in a high-resolution generated image. 
Patient reports from the MIMIC-CXR dataset were utilized to conditionally create chest X-rays \cite{johnson2019mimic}. Each report corresponds to a real chest X-ray image from an imaging study. To avoid unintentional loss of information, selected reports must meet RoentGen's token length limit. These reports were then processed through RoentGen, producing synthetic chest X-ray images at a resolution of 512x512 with 75 inference steps.

\subsubsection{Real Stimuli}
 
The MIMIC-CXR dataset contains 227,835 imaging studies with 30 reports being randomly selected from this set. As each synthetic chest X-ray was generated from a report in the MIMIC-CXR dataset, each synthetic chest X-ray has a corresponding real X-ray with the same report content. These corresponding X-rays were the ones selected to be used in the study. 

\subsection{Analysis of Population}
The population under analysis comprises 5 women and 11 men, totaling 16 individuals. The distribution of years of experience is as follows: 2 individuals have 0-5 years of experience, 6 have 6-10 years, 4 have 10-20 years, and 3 have over 20 years of experience. The group also includes various specialties including: body imaging, abdominal imaging, cardiothoracic radiology, interventional radiology, neuroradiology, musculoskeletal radiology, and thoracic radiology.

\begin{table*}[ht]
    \centering
    \caption{Saccades Amplitude Statistics for Real and Fake image ($Visual ^\circ$) }
    \begin{tabular}{lccccc}
        \toprule
        Image Type & Min & Max & Mean & Median & Std Dev \\
        \midrule
        Fake & 0.0269 & 86.1114 & 5.7080 & 4.3961 & 4.5628 \\
        Real & 0.0210 & 83.6801 & 5.9254 & 4.6179 & 4.8184 \\
        \bottomrule
    \end{tabular}
    \label{tab:scale_amplitudes}
\end{table*}

\subsection{Analysis of Visual Behavior Patterns}

In order to verify our hypothesis (i.e. Radiologist exhibit different visual behaviors when viewing real X-Ray images vs AI Generated X-Ray images), we analyze the statistical patterns of the scanpath data collected during the experiments. This includes the saccadic length, direction, the joint distribution of the saccadic length and direction, as well as a temporal analysis of the first fixation, the last fixation and the fixations with the longest and shortest duration during the viewing.    

\subsubsection{Saccadic Amplitudes}

Table \ref{tab:scale_amplitudes} represents the descriptive statistics of the saccadic amplitudes distribution.  

The data suggests that radiologists exhibit slightly different eye movement patterns when viewing real versus fake X-ray images. The larger maximum saccadic amplitudes for fake images might indicate more extensive searching or difficulty in interpreting these images. The lower mean and median saccadic amplitudes for fake images suggest smaller, more cautious eye movements, potentially reflecting uncertainty or unfamiliarity. The lower standard deviation for fake images indicates more consistent eye-movement patterns, which could be due to a more systematic approach to scanning these images.
These conclusions are further supported by the histograms shown in Fig.\ref{fig:scale_amplitudes}(a) and Fig.\ref{fig:scale_amplitudes}(b), which provide a visual representation of the saccadic amplitude distributions for both real and fake images and show that the images are more heavily distributed on the left.

\subsubsection{Saccadic Direction}

The polar histograms in Fig.\ref{fig:scale_amplitudes}(c) and Fig.\ref{fig:scale_amplitudes}(d) illustrate the distributions of saccadic directions for both real and fake images. While the overall patterns are similar, there are subtle differences between the two distributions.
Both histograms reveal a high concentration of saccades in the horizontal and vertical directions. This phenomenon can be attributed to several factors, including anatomical constraints and evolutionary adaptations. The predominance of horizontal and vertical saccades reflects the natural tendencies of human eye movements, which are optimized for scanning and interpreting visual information in these directions.

\subsubsection{Joint Distribution of Saccadic Amplitudes and Direction}

\begin{figure*}[ht]
    \centering
    \includegraphics[width=1.0\textwidth]{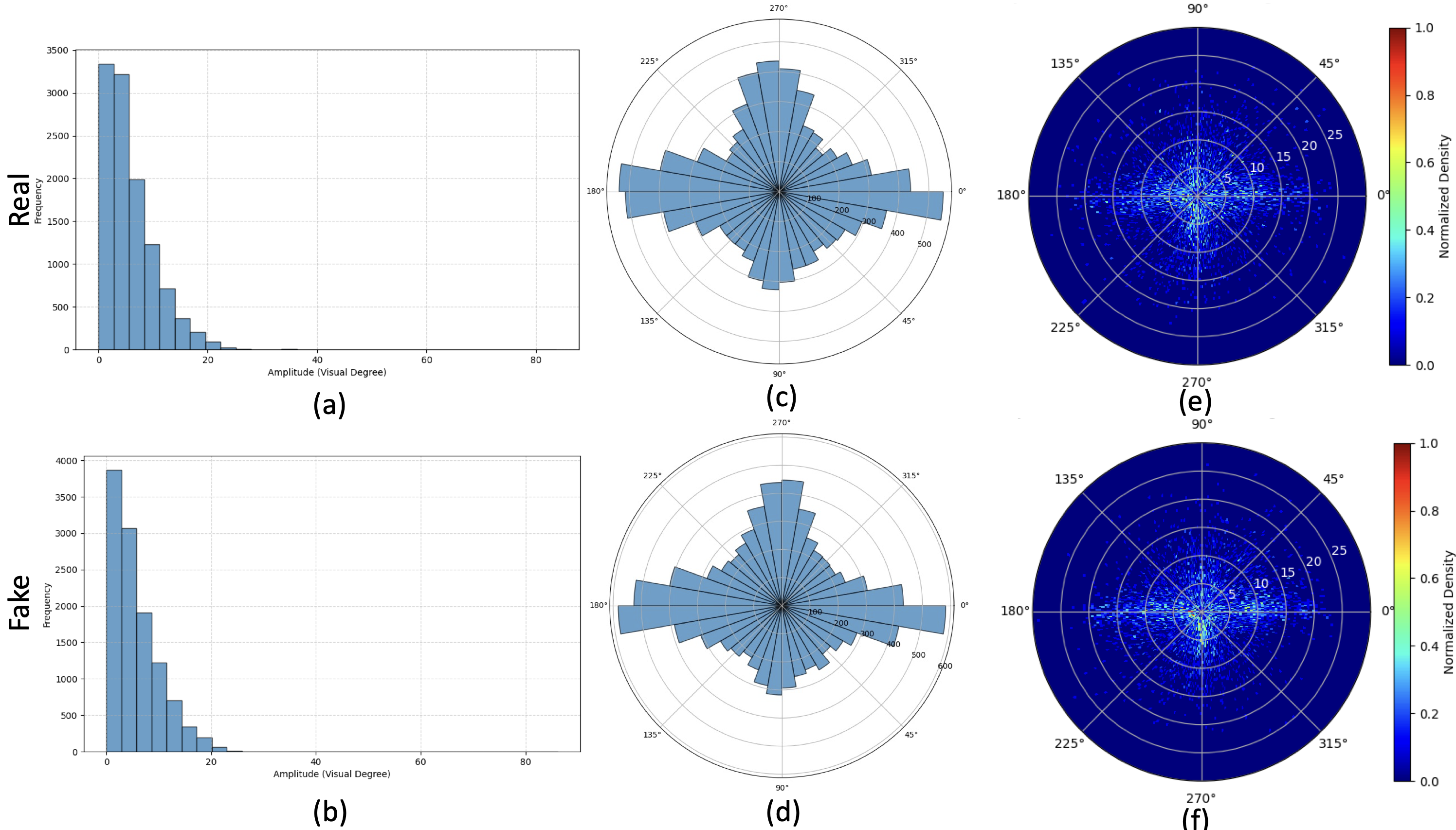} 
    \caption{(a+b) Comparision of saccadic length distribution, (c+d) Comparision of saccadic direction distribution, (a+b) Comparision of the joint saccadic length and direction distribution.}
    \label{fig:scale_amplitudes}
\end{figure*}

Fig.\ref{fig:scale_amplitudes}(e) and Fig.\ref{fig:scale_amplitudes}(f) present the joint distributions of saccadic amplitudes and directions. The heatmap distributions corroborate the previous findings and are consistent with the patterns identified in prior research \cite{LeMeur2015} on common eye-movement patterns for multiple image domains \cite{LeMeur2020}. 

These visualizations provide a comprehensive view of how saccadic amplitudes and directions interact, further validating the observed trends in radiologists' eye movements when interpreting real and fake X-ray images.

\subsubsection{Temporal Analysis}
Fixation duration is a crucial characteristic when analyzing visual behavior patterns. To this end, we extracted statistical descriptive aggregations for the first and last fixations during the viewing experience, as well as the fixations with the longest and shortest viewing times. These metrics, presented in Table.\ref{tab:fixation_time} provide valuable insights into how viewers engage with the images and can help identify key areas of interest and patterns in visual attention.

\begin{table*}[ht]
    \centering
    
    \caption{Fixation duration Statistics for Real and Fake images Maps}
    \begin{tabular}{llccccc}
        \toprule
        Image Type & Fixation Type & Min Time & Max Time & Mean Time & Median Time & Std Dev Time 
       \\
        \midrule
        Fake & First        & 12.0  & 1030.0  & 279.62314  & 268.0  & 98.07551  \\
        Fake & Last         & 32.0  & 1186.0  & 283.9093   & 250.0  & 177.01553 \\
        Fake & Longest    & 168.0 & 2284.0  & 575.86786  & 501.0  & 248.73659 \\
        Fake & Shortest     & 12.0  & 260.0   & 102.41724  & 104.0  & 52.94234  \\
        \midrule

        Real & First        & 14.0  & 664.0   & 277.23605  & 270.0  & 103.05347  \\
        Real & Last         & 30.0  & 1176.0  & 294.81897  & 266.0  & 181.40292  \\
        Real & Longest    & 264.0 & 3252.0  & 559.27826  & 488.0  & 263.99258  \\
        Real & Shortest     & 4.0   & 532.0   & 108.43421  & 106.0  & 63.561237  \\
        \bottomrule
    \end{tabular}
    \label{tab:fixation_time}
\end{table*}

\begin{itemize}
    \item \textbf{Analysis of First Fixation distribution}:
        The first fixation times for both real and fake images are quite similar, indicating that initial attention is captured similarly regardless of the image type. The slightly higher minimum and maximum values for fake images suggest a broader range of initial fixation durations, but the mean and median values are nearly identical, showing consistent initial engagement.

    \item \textbf{Analysis of Last Fixation distribution}: 
        The last fixation times are also similar between real and fake images, suggesting that the final moments of viewing are consistent across both types. The slightly higher mean and median values for real images indicate a marginally longer final engagement, but the overall distribution is comparable.

    \item \textbf{Analysis of Longest Fixation distribution}: 
        The longest fixation times are slightly higher for fake images, indicating that certain areas in fake images may require more attention, and by extension cognitive load. The broader range and higher standard deviation for real images suggest more variability in the longest fixations, possibly reflecting a wider range of areas requiring detailed examination at more consistent mental effort.

      \item \textbf{Analysis of Shortest Fixation distribution}: 
        The shortest fixation times are slightly higher for real images, suggesting that quick glances are more common in real images. The broader range and higher standard deviation for real images indicate more variability in the shortest fixations, possibly reflecting a wider range of areas that capture brief attention.

\end{itemize}

The fixation durations for real and fake images show only minor differences, with real images generally exhibiting slightly higher variability in both the longest and shortest fixations. The first and last fixations are nearly identical, indicating consistent initial and final attention. The longest fixations are slightly longer for fake images, suggesting increased scrutiny, while the shortest fixations are slightly longer for real images, indicating more frequent quick glances. These patterns provide insights into how radiologists engage with real and fake images during their viewing experience, highlighting subtle differences in visual behavior.

\subsection{Spatial Analysis of Fixation of Interest}

\begin{figure*}[ht]
    \centering
    \includegraphics[width=1.0\textwidth]{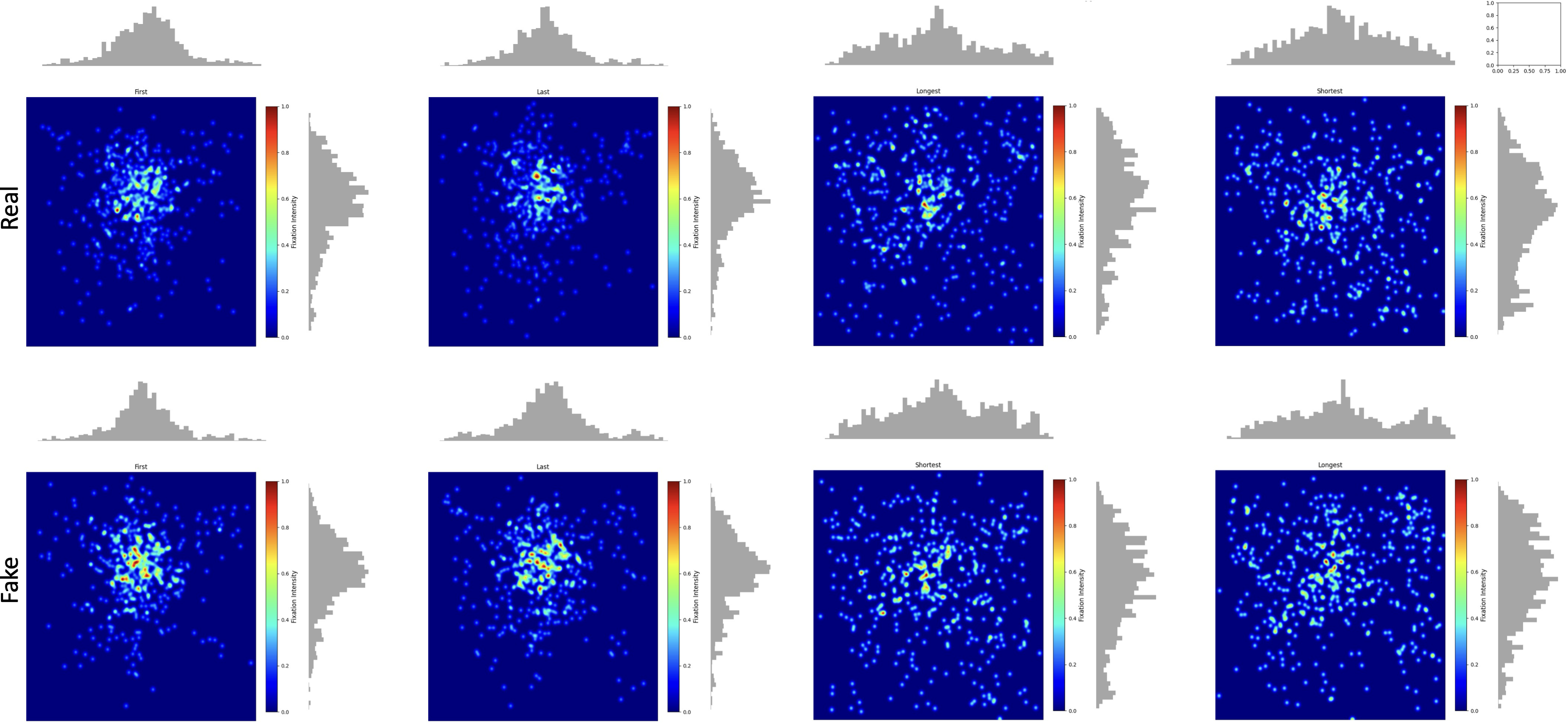} 
    \caption{Visualization of Bias Maps between real and fake in the order of: First, Last, Longest, Shortest.}
    \label{fig:biases}
\end{figure*}

\begin{table*}[ht]
    \centering
    \caption{Saliency Metrics comparing Real and Fake Bias Maps \cite{metrics} }
    \begin{tabular}{lccc}
        \toprule
        Condition & Correlation Coefficient & KL Divergence  & Similarity Metric \\
        \midrule
        First     & 0.5158 & 7.0446 & 0.4286 \\
        Last      & 0.4777 & 6.8679 & 0.4104 \\
        Longest & 0.1938 & 9.7120 & 0.2839 \\
        Shortest  & 0.2044 & 10.2255 & 0.2860 \\
        \bottomrule
    \end{tabular}
    \label{tab:metrics}
\end{table*}

In addition to the previous results obtained through bias saliency maps. Table. \ref{tab:metrics} presents a comparative study of viewing patterns biases on fake and real as represented in Fig.\ref{fig:biases}, we used several saliency distribution-based metrics: Correlation Coefficient, KL Divergence, and Similarity Metric\cite{metrics}. Each metric provides insights into how closely the viewing patterns for fake images align with those for real images under different conditions (First, Last, Longest, and Shortest fixations).

\begin{itemize}
    \item \textbf{CC: }The correlation coefficient measures the strength and direction of the linear relationship between the viewing patterns of real and fake images. Higher values indicate a stronger correlation. The first and last fixations have relatively higher correlation coefficients, suggesting that the initial and final viewing patterns for fake images are more similar to those for real images, This implies that radiologists' initial attention and final moments of viewing is captured in a similar manner for both image types.     
    The longest and shortest fixations have lower correlation coefficients, indicating weaker alignment in these conditions.  This implies that radiologists may exhibit different patterns of prolonged attention when viewing fake images, as well as  different patterns of brief attention when viewing fake images.

    \item \textbf{KLD}: KL Divergence measures the difference between two probability distributions. Lower values indicate that the distributions are more similar. The first and last fixations have lower KL Divergence values, suggesting that the saliency distributions for fake images are more similar to those for real images in these conditions. The longest and shortest fixations have higher KL Divergence values, indicating greater differences in the saliency distributions.
    Results of the Longest fixation indicates a larger difference between the saliency distributions of real and fake images.This suggests that the patterns of prolonged attention differ more significantly between the two image types.
    The results on the Shortest fixation also indicates a larger difference between the saliency distributions of real and fake images. This suggests that the patterns of brief attention differ more significantly between the two image types

    \item \textbf{Similarity: } The similarity metric \cite{metrics} quantifies the overall similarity between the saliency maps of real and fake images. Higher values indicate greater similarity. The first and last fixations have higher similarity metrics, indicating that the saliency maps for fake images are more similar to those for real images in these conditions. The longest and shortest fixations have lower similarity, suggesting less similarity in these conditions. 
    This suggests that the initial and final attention patterns are somewhat similar across both image types.
    On the other hand the results on Longest and Shortest fixation imply that hat prolonged and brief transient attention patterns differ more significantly between the two image types.
\end{itemize}
The analysis reveals that the viewing patterns for fake images are more closely aligned with those for real images during the first and last fixations. This implies that initial and final attention patterns are more consistent between real and fake images. 
In contrast, the longest and shortest fixations show weaker alignment, indicating greater differences in viewing patterns during these conditions.
The same shift can be notices on prolonged attention. 
This concludes that while initial and final attention align on both the temporal and spatial dimensions, viewing behavior shows larger discrepancies at the extremes of the temporal space, affecting both spatial and temporal dimensions, this can be visualized through the Bias maps in Figure. \ref{fig:biases}, where we notice similar distributions, which slightly differ for the Longest and Shortest fixations. The maps also unveil that while the First and Last fixations are more congruent in distribution, the Longest and Shortest are more dispersed on the spatial domain.

\section{Privacy and Ethics Statement}  
This study adheres to ethical guidelines for data privacy and patient protection, using AI-generated medical images solely for research purposes. We acknowledge potential risks, including bias in synthetic images and their misuse in clinical decision-making, emphasizing the need for careful integration into healthcare.  

\section{Conclusion}
This study provides a detailed analysis of radiologists' visual behavior when interpreting real and fake X-ray images. By examining various fixation metrics, including first, last, longest, and shortest fixations, we identified subtle, but significant differences in how radiologists engage with these images. The initial and final attention patterns were found to be consistent across both real and fake images, suggesting similar engagement at the beginning and end of the viewing process. However, notable discrepancies were observed in the extremes of the temporal space, particularly in the longest and shortest fixations, indicating that fake images may require more scrutiny or exhibit distinct visual characteristics that require different viewing strategies.
The saliency metrics and the statistical analysis of the gaze patterns further supported these findings, showing that while there is a moderate alignment in initial and final fixations, the alignment weakens significantly for the longest and shortest fixations. This suggests that radiologists may need to adopt different approaches when interpreting fake images, potentially due to increased uncertainty or unfamiliarity with these images.
In general, these findings provide valuable insight into how radiologists' visual behavior differs when interpreting real versus fake images, highlighting areas where fake images may require more scrutiny or exhibit distinct visual characteristics.

\small
\bibliographystyle{IEEEbib}
\bibliography{strings}

\begin{thebibliography}{10}

\bibitem{Bercovich2018}
Eyal Bercovich and Marcia~C. Javitt,
\newblock ``Medical imaging: From roentgen to the digital revolution, and beyond,''
\newblock {\em Rambam Maimonides Medical Journal}, vol. 9, no. 4, pp. e0034, Oct. 2018.

\bibitem{Atabo2019}
Shaibu~Mohammed Atabo and Abubakar~Abubakar Umar,
\newblock ``A review of imaging techniques in scientific research/clinical diagnosis,''
\newblock {\em MOJ Anatomy \& Physiology}, vol. 6, no. 5, Oct. 2019.

\bibitem{Hussain2022}
Shah Hussain, Iqra Mubeen, Niamat Ullah, Syed Shahab Ud~Din Shah, Bakhtawar~Abduljalil Khan, Muhammad Zahoor, Riaz Ullah, Farhat~Ali Khan, and Mujeeb~A. Sultan,
\newblock ``Modern diagnostic imaging technique applications and risk factors in the medical field: A review,''
\newblock {\em BioMed Research International}, vol. 2022, no. 1, Jan. 2022.

\bibitem{Najjar2023}
Reabal Najjar,
\newblock ``Redefining radiology: A review of artificial intelligence integration in medical imaging,''
\newblock {\em Diagnostics}, vol. 13, no. 17, pp. 2760, Aug. 2023.

\bibitem{rad2023}
Tajamul Rashid~Wani ~ and Muteeb Showket~Reshi ~,
\newblock ``Revolutionizing radiology: Exploring applications and advancements in ai for imaging diagnostics,''
\newblock {\em International Journal For Multidisciplinary Research}, vol. 5, no. 6, Dec. 2023.

\bibitem{GANs}
Ian Goodfellow, Jean Pouget-Abadie, Mehdi Mirza, Bing Xu, David Warde-Farley, Sherjil Ozair, Aaron Courville, and Yoshua Bengio,
\newblock ``Generative adversarial nets,''
\newblock in {\em Advances in Neural Information Processing Systems}, Z.~Ghahramani, M.~Welling, C.~Cortes, N.~Lawrence, and K.Q. Weinberger, Eds. 2014, vol.~27, Curran Associates, Inc.

\bibitem{stable_diffusion}
Robin Rombach, A.~Blattmann, Dominik Lorenz, Patrick Esser, and Bj{\"o}rn Ommer,
\newblock ``High-resolution image synthesis with latent diffusion models,''
\newblock {\em 2022 IEEE/CVF Conference on Computer Vision and Pattern Recognition (CVPR)}, pp. 10674--10685, 2021.

\bibitem{synt_med_gen1}
John~R. McNulty, Lee Kho, Alexandria~L. Case, Charlie Fornaca, Drew Johnston, David Slater, Joshua~M. Abzug, and Sybil~A. Russell,
\newblock ``Synthetic medical imaging generation with generative adversarial networks for plain radiographs,'' 2024.

\bibitem{synt_med_gen2}
Loc~X. Nguyen, Pyae Sone~Aung, Huy~Q. Le, Seong-Bae Park, and Choong~Seon Hong,
\newblock ``A new chapter for medical image generation: The stable diffusion method,''
\newblock in {\em 2023 International Conference on Information Networking (ICOIN)}. Jan. 2023, p. 483–486, IEEE.

\bibitem{Islam2024}
Showrov Islam, Md.~Tarek Aziz, Hadiur~Rahman Nabil, Jamin~Rahman Jim, M.~F. Mridha, Md.~Mohsin Kabir, Nobuyoshi Asai, and Jungpil Shin,
\newblock ``Generative adversarial networks (gans) in medical imaging: Advancements, applications, and challenges,''
\newblock {\em IEEE Access}, vol. 12, pp. 35728–35753, 2024.

\bibitem{brain_img_syth}
Firoozeh~Shomal Zadeh, Sevda Molani, Maysam Orouskhani, Marziyeh Rezaei, Mehrzad Shafiei, and Hossein Abbasi,
\newblock ``Generative adversarial networks for brain images synthesis: A review,'' 2023.

\bibitem{Musalamadugu2023}
Tanmai~Sree Musalamadugu and Hemachandran Kannan,
\newblock ``Generative ai for medical imaging analysis and applications,''
\newblock {\em Future Medicine AI}, Sept. 2023.

\bibitem{Kelkar2023}
Varun~A. Kelkar, Dimitrios~S. Gotsis, Frank~J. Brooks, Prabhat KC, Kyle~J. Myers, Rongping Zeng, and Mark~A. Anastasio,
\newblock ``Assessing the ability of generative adversarial networks to learn canonical medical image statistics,''
\newblock {\em IEEE Transactions on Medical Imaging}, vol. 42, no. 6, pp. 1799–1808, June 2023.

\bibitem{ETML_med}
Sahar Moradizeyveh, Mehnaz Tabassum, Sidong Liu, Robert~Ahadizad Newport, Amin Beheshti, and Antonio Di~Ieva,
\newblock ``When eye-tracking meets machine learning: A systematic review on applications in medical image analysis,'' 2024.

\bibitem{Trkan2007HumanEL}
Mehmet T{\"u}rkan, Montse Pard{\`a}s, and A.~Enis Çetin,
\newblock ``Human eye localization using edge projections,''
\newblock in {\em International Conference on Computer Vision Theory and Applications}, 2007.

\bibitem{Kerkouri2022}
Mohamed~Amine Kerkouri, Marouane Tliba, Aladine Chetouani, and Alessandro Bruno,
\newblock ``A domain adaptive deep learning solution for scanpath prediction of paintings,''
\newblock in {\em International Conference on Content-based Multimedia Indexing}. Sept. 2022, CBMI 2022, p. 57–63, ACM.

\bibitem{Tliba_2022_CVPR}
Marouane Tliba, Mohamed~Amine Kerkouri, Aladine Chetouani, and Alessandro Bruno,
\newblock ``Self supervised scanpath prediction framework for painting images,''
\newblock in {\em Proceedings of the IEEE/CVF Conference on Computer Vision and Pattern Recognition (CVPR) Workshops}, June 2022, pp. 1539--1548.

\bibitem{1334654}
A.M. Bagci, R.~Ansari, A.~Khokhar, and E.~Cetin,
\newblock ``Eye tracking using markov models,''
\newblock in {\em Proceedings of the 17th International Conference on Pattern Recognition, 2004. ICPR 2004.}, 2004, vol.~3, pp. 818--821 Vol.3.

\bibitem{Wu2019}
Chia-Chien Wu and Jeremy~M. Wolfe,
\newblock ``Eye movements in medical image perception: A selective review of past, present and future,''
\newblock {\em Vision}, vol. 3, no. 2, pp. 32, June 2019.

\bibitem{Lou2021}
Jianxun Lou, Xin Zhao, Philippa Young, Richard White, and Hantao Liu,
\newblock ``Study of saccadic eye movements in diagnostic imaging,''
\newblock in {\em 2021 IEEE International Conference on Image Processing (ICIP)}. Sept. 2021, p. 1474–1478, IEEE.

\bibitem{Leveque2021}
Lucie Leveque, Philippa Young, and Hantao Liu,
\newblock ``Studying the gaze patterns of expert radiologists in screening mammography: A case study with breast test wales,''
\newblock in {\em 2020 28th European Signal Processing Conference (EUSIPCO)}. Jan. 2021, p. 1249–1253, IEEE.

\bibitem{K2024}
Aiswariya~Milan K, Amudha J, and George Ghinea,
\newblock ``Automated insight tool: Analyzing eye tracking data of expert and novice radiologists during optic disc detection task,''
\newblock in {\em Proceedings of the 2024 Symposium on Eye Tracking Research and Applications}. June 2024, ETRA ’24, p. 1–7, ACM.

\bibitem{Pershin2023}
Ilya Pershin, Tamerlan Mustafaev, and Bulat Ibragimov,
\newblock ``Contrastive learning approach to predict radiologist’s error based on gaze data,''
\newblock in {\em 2023 IEEE Congress on Evolutionary Computation (CEC)}. July 2023, p. 1–6, IEEE.

\bibitem{9051780}
Diederik~P. Kingma and Max Welling,
\newblock ``An introduction to variational autoencoders,''
\newblock {\em Found. Trends Mach. Learn.}, vol. 12, pp. 307--392, 2019.

\bibitem{Zingarini2024}
G.~Zingarini, D.~Cozzolino, R.~Corvi, G.~Poggi, and L.~Verdoliva,
\newblock ``M3dsynth: A dataset of medical 3d images with ai-generated local manipulations,''
\newblock in {\em ICASSP 2024 - 2024 IEEE International Conference on Acoustics, Speech and Signal Processing (ICASSP)}. Apr. 2024, p. 13176–13180, IEEE.

\bibitem{helmholtz1924treatise}
H.~v. Helmholtz and J.~P.~C. Southall,
\newblock ``Helmholtz's treatise on physiological optics, vol 1 (trans. from the 3rd german ed.).,''
\newblock 1924.

\bibitem{Yarbus1967}
A.~L. Yarbus,
\newblock {\em Eye Movements and Vision},
\newblock Plenum. New York., 1967.

\bibitem{Kosslyn1994}
Stephen~M. Kosslyn,
\newblock {\em {Image And Brain: The Resolution of the Imagery Debate}},
\newblock The MIT Press, 06 1994.

\bibitem{Nodine1987}
C~F Nodine and H~L Kundel,
\newblock ``Using eye movements to study visual search and to improve tumor detection.,''
\newblock {\em RadioGraphics}, vol. 7, no. 6, pp. 1241–1250, Nov. 1987.

\bibitem{KUNDEL1990}
Harold~L. Kundel, Calvin~F. Nodine, and Elizabeth~A. Krupinski,
\newblock ``Computer-displayed eye position as a visual aid to pulmonary nodule interpretation,''
\newblock {\em Investigative Radiology}, vol. 25, no. 8, pp. 890–896, Aug. 1990.

\bibitem{Litchfield2010}
Damien Litchfield, Linden~J. Ball, Tim Donovan, David~J. Manning, and Trevor Crawford,
\newblock ``Viewing another person’s eye movements improves identification of pulmonary nodules in chest x-ray inspection.,''
\newblock {\em Journal of Experimental Psychology: Applied}, vol. 16, no. 3, pp. 251–262, Sept. 2010.

\bibitem{Borg2018PreliminaryEU}
Lindsay~K Borg, T~Kyle Harrison, Alex Kou, Edward~R. Mariano, Ankeet~D. Udani, T~Edward Kim, Cynthia Shum, and Steven~K. Howard,
\newblock ``Preliminary experience using eye‐tracking technology to differentiate novice and expert image interpretation for ultrasound‐guided regional anesthesia,''
\newblock {\em Journal of Ultrasound in Medicine}, vol. 37, 2018.

\bibitem{Bluethgen2024}
Christian Bluethgen, Pierre Chambon, Jean-Benoit Delbrouck, Rogier van~der Sluijs, Ma{\l}gorzata Po{\l}acin, Juan~Manuel Zambrano~Chaves, Tanishq~Mathew Abraham, Shivanshu Purohit, Curtis~P. Langlotz, and Akshay~S. Chaudhari,
\newblock ``A vision--language foundation model for the generation of realistic chest x-ray images,''
\newblock {\em Nature Biomedical Engineering}, Aug 2024.

\bibitem{Nie2018}
Dong Nie, Roger Trullo, Jun Lian, Li~Wang, Caroline Petitjean, Su~Ruan, Qian Wang, and Dinggang Shen,
\newblock ``Medical image synthesis with deep convolutional adversarial networks,''
\newblock {\em IEEE Transactions on Biomedical Engineering}, vol. 65, no. 12, pp. 2720–2730, Dec. 2018.

\bibitem{Lee2020}
Jung~Hwan Lee, In~Ho Han, Dong~Hwan Kim, Seunghan Yu, In~Sook Lee, You~Seon Song, Seongsu Joo, Cheng-Bin Jin, and Hakil Kim,
\newblock ``Spine computed tomography to magnetic resonance image synthesis using generative adversarial networks: A preliminary study,''
\newblock {\em Journal of Korean Neurosurgical Society}, vol. 63, no. 3, pp. 386–396, May 2020.

\bibitem{Ciano2021}
Giorgio Ciano, Paolo Andreini, Tommaso Mazzierli, Monica Bianchini, and Franco Scarselli,
\newblock ``A multi-stage gan for multi-organ chest x-ray image generation and segmentation,''
\newblock {\em Mathematics}, vol. 9, no. 22, pp. 2896, Nov. 2021.

\bibitem{diff_anat}
Nicholas Konz, Yuwen Chen, Haoyu Dong, and Maciej~A. Mazurowski,
\newblock ``Anatomically-controllable medical image generation with segmentation-guided diffusion models,'' 2024.

\bibitem{med_diff}
Firas Khader, Gustav Mueller-Franzes, Soroosh~Tayebi Arasteh, Tianyu Han, Christoph Haarburger, Maximilian Schulze-Hagen, Philipp Schad, Sandy Engelhardt, Bettina Baessler, Sebastian Foersch, Johannes Stegmaier, Christiane Kuhl, Sven Nebelung, Jakob~Nikolas Kather, and Daniel Truhn,
\newblock ``Medical diffusion: Denoising diffusion probabilistic models for 3d medical image generation,'' 2022.

\bibitem{Steinfeld2023}
Nili Steinfeld,
\newblock ``How do users examine online messages to determine if they are credible? an eye-tracking study of digital literacy, visual attention to metadata, and success in misinformation identification,''
\newblock {\em Social Media + Society}, vol. 9, no. 3, July 2023.

\bibitem{huang2024analysishumanperceptiondistinguishing}
Jin Huang, Subhadra Gopalakrishnan, Trisha Mittal, Jake Zuena, and Jaclyn Pytlarz,
\newblock ``Analysis of human perception in distinguishing real and ai-generated faces: An eye-tracking based study,'' 2024.

\bibitem{cornelissen2002eyelink}
Frans~W Cornelissen, Enno~M Peters, and John Palmer,
\newblock ``The eyelink toolbox: eye tracking with matlab and the psychophysics toolbox,''
\newblock {\em Behavior Research Methods, Instruments, \& Computers}, vol. 34, no. 4, pp. 613--617, 2002.

\bibitem{chambon2022roentgen}
Pierre Chambon, Christian Bluethgen, Jean-Benoit Delbrouck, Rogier Van~der Sluijs, Ma{\l}gorzata Po{\l}acin, Juan Manuel~Zambrano Chaves, Tanishq~Mathew Abraham, Shivanshu Purohit, Curtis~P Langlotz, and Akshay Chaudhari,
\newblock ``Roentgen: vision-language foundation model for chest x-ray generation,''
\newblock {\em arXiv preprint arXiv:2211.12737}, 2022.

\bibitem{johnson2019mimic}
Alistair~EW Johnson, Tom~J Pollard, Seth~J Berkowitz, Nathaniel~R Greenbaum, Matthew~P Lungren, Chih-ying Deng, Roger~G Mark, and Steven Horng,
\newblock ``Mimic-cxr, a de-identified publicly available database of chest radiographs with free-text reports,''
\newblock {\em Scientific data}, vol. 6, no. 1, pp. 317, 2019.

\bibitem{LeMeur2015}
Olivier Le~Meur and Zhi Liu,
\newblock ``Saccadic model of eye movements for free-viewing condition,''
\newblock {\em Vision Research}, vol. 116, pp. 152–164, Nov. 2015.

\bibitem{LeMeur2020}
Olivier Le~Meur, Tugdual Le~Pen, and Rémi Cozot,
\newblock ``Can we accurately predict where we look at paintings?,''
\newblock {\em PLOS ONE}, vol. 15, no. 10, pp. e0239980, Oct. 2020.

\bibitem{metrics}
Matthias K{\"u}mmerer, Thomas S.~A. Wallis, and Matthias Bethge,
\newblock ``Saliency benchmarking made easy: Separating models, maps and metrics,''
\newblock in {\em Computer Vision – {ECCV} 2018}, Vittorio Ferrari, Martial Hebert, Cristian Sminchisescu, and Yair Weiss, Eds. Lecture Notes in Computer Science, pp. 798--814, Springer International Publishing.

\end{thebibliography}

\end{document}